\documentclass[11pt]{article}

\usepackage[preprint]{acl}

\usepackage{times}
\usepackage{latexsym}

\usepackage[T1]{fontenc}

\usepackage[utf8]{inputenc}

\usepackage{microtype}

\usepackage{inconsolata}

\usepackage{graphicx}
\usepackage{stfloats}
\usepackage{placeins}

\usepackage{booktabs}
\usepackage{multirow}
\usepackage{amsmath,amssymb}
\usepackage{xcolor}
\usepackage{enumitem}
\usepackage{listings}
\lstdefinestyle{prompt}{
  basicstyle=\small\ttfamily,
  columns=fullflexible,
  keepspaces=true,
  breaklines=true,
  breakatwhitespace=true,
  showstringspaces=false,
  frame=single,
  framerule=0.3pt,
  rulecolor=\color{gray},
  backgroundcolor=\color{gray!3},
  xleftmargin=3pt,
  xrightmargin=3pt,
  framexleftmargin=3pt,
  framexrightmargin=3pt,
  aboveskip=6pt,
  belowskip=6pt
}
\usepackage[normalem]{ulem}
\usepackage{xcolor}

\definecolor{scopecolor}{HTML}{0072B2}
\definecolor{termcolor}{HTML}{8E44AD}
\definecolor{prefcolor}{HTML}{B35806}
\newcommand{\scopesout}[1]{\textcolor{scopecolor}{\sout{#1}}}
\newcommand{\termsout}[1]{\textcolor{termcolor}{\sout{#1}}}
\newcommand{\prefsout}[1]{\textcolor{prefcolor}{\sout{#1}}}

\newcommand{\tba}[1]{\textcolor{red}{[\ifx\relax#1\relax TBA\else #1\fi]}}

\definecolor{hz}{HTML}{7E22CE}

\newcommand{\sys}{\textsc{UnderSpec-DR}}
\newcommand{\tulu}{DR~Tulu-8B}
\newcommand{\tulugpt}{GPT-5.4 with tools}
\newcommand{\tongyi}{Tongyi-DR}
\newcommand{\rqa}{ResearchQA}
\newcommand{\sqa}{ScholarQA-Bench}
\newcommand{\drb}{DeepResearch Bench}

\newcommand{\qS}{q_{\textsc{s}}}
\newcommand{\qT}{q_{\textsc{t}}}
\newcommand{\qP}{q_{\textsc{p}}}
\newcommand{\qU}{q_{\textsc{u}}}
\newcommand{\sig}[1]{\textsuperscript{#1}}

\title{
\includegraphics[height=1.2em]{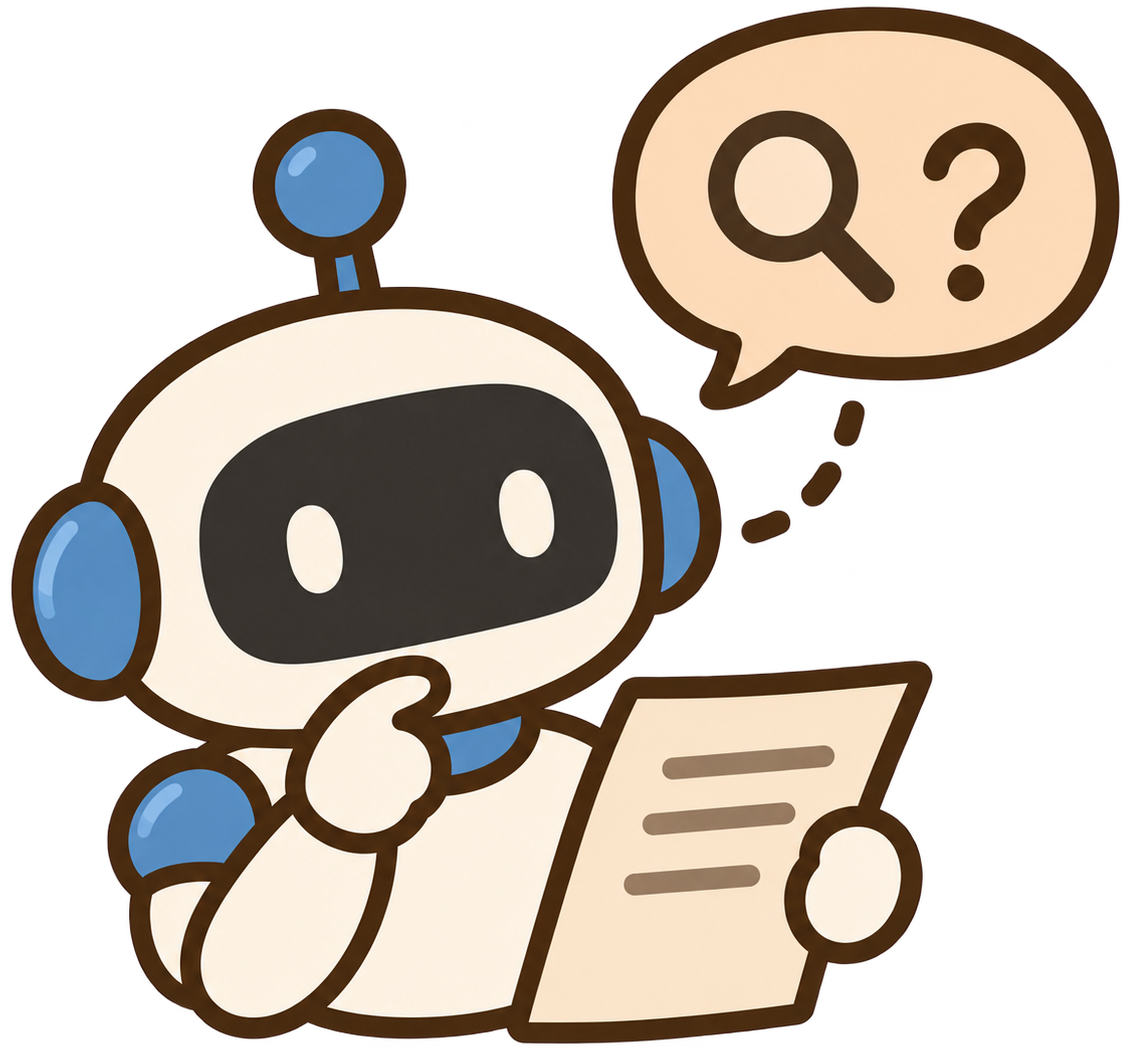} How Specific Do I Need to Be?\\
Effects of User Query Underspecification on Deep Research Agents
}
\author{
  \textbf{Yi Wen\textsuperscript{1,}}\thanks{Equal contribution.},
  \textbf{Xiangjue Dong\textsuperscript{2,*}},
  \textbf{Bingxuan Li\textsuperscript{3}},
  \textbf{Hangxiao Zhu\textsuperscript{2}} \\
  \textbf{Leo Zhicheng Liu\textsuperscript{1}},
  \textbf{Yu Zhang\textsuperscript{2,}}\thanks{Corresponding authors.},
  \textbf{James Caverlee\textsuperscript{2,\ensuremath{\dagger}}} \\[4pt]
  \textsuperscript{1}\,University of Maryland, College Park\quad
  \textsuperscript{2}\,Texas A\&M University \\
  \textsuperscript{3}\,University of Illinois, Urbana-Champaign
}

\begin{document}
\raggedbottom
\begin{NoHyper}
\maketitle
\end{NoHyper}
\begin{abstract}
Deep research agents (DRAs) conduct long-horizon reasoning, search, and literature synthesis to provide long-form answers to complex, domain-specific questions. These agents are typically trained and evaluated on fully specified user queries. However, real-world user queries can often be underspecified. We introduce \sys{}, a framework for systematically evaluating how underspecified queries affect agents' final answer quality and intermediate search behavior. 
Across three DRAs we evaluated, underspecification substantially reduces how well DRA's answers fulfill users’ information needs, even though they remain grounded in cited evidence. Our analysis of intermediate search behavior suggests a possible explanation: DRAs conduct broader searches that draw less on the original user query but rely more on their learned prior during training, which can be misaligned with users' intent. Our findings reveal the vulnerability of current DRAs to underspecified user queries and inform future efforts to build DRAs that proactively recognize when additional user input is needed to reliably complete deep research tasks.


\end{abstract}

\section{Introduction}

Deep research agents (DRAs), which autonomously plan multi-step searches, retrieve documents, read and synthesize them into long-form reports with citations, have been deployed in both commercial \citep{openai2025deepresearch,google2025deepresearch} and open-source systems \citep{shao2025drtulu,tongyi2025deepresearch}. Existing benchmarks \citep{mialon2024gaia,wei2025browsecomp,asai2024openscholar,li2025researchqa,du2026deepresearchbench} for evaluating DRAs carefully construct fully-specified user queries to enable consistent assessment of the agent's deep research capability. Deep research requires precise questions, but in practice, users often cannot formulate their query as required. They may not yet have a clear understanding of their information needs, or they may lack the domain knowledge needed to express those needs precisely~\citep{taylor1968question,belkin1982ask}. This creates a fundamental tension: DRAs are designed to answer questions that users cannot easily answer themselves, yet they are developed and evaluated on the assumption that users can state those questions perfectly. This tension motivates the central question of our work: \textbf{\emph{Can DRAs handle cases when users cannot fully articulate those questions in the first place?}}

 Prior work has examined how underspecified user queries affect performance in both standalone LLMs \citep{yang-etal-2026-prompts} and agentic settings \citep{cao2026clarifyOA, vijayvargiya2026ambigswe}, and has developed methods to evaluate and improve their ability to resolve ambiguity \citep{sun2026training}. There are verifiable outcomes under these settings, such as whether a stated requirement is met or whether the code generated passed the test cases. However, DRAs produce long, open-ended reports that are difficult to verify, which makes it hard to measure how much underspecification harms the final answer DRAs generated and whether the they resolved ambiguity. Moreover, unlike standalone LLMs, DRAs interact with external tools over long-horizon task execution. Inspecting this intermediate process might provide an additional lens for understanding the effect of user query underspecification on DRA.


\begin{figure*}[!ht]
\centering
\includegraphics[width=\textwidth]{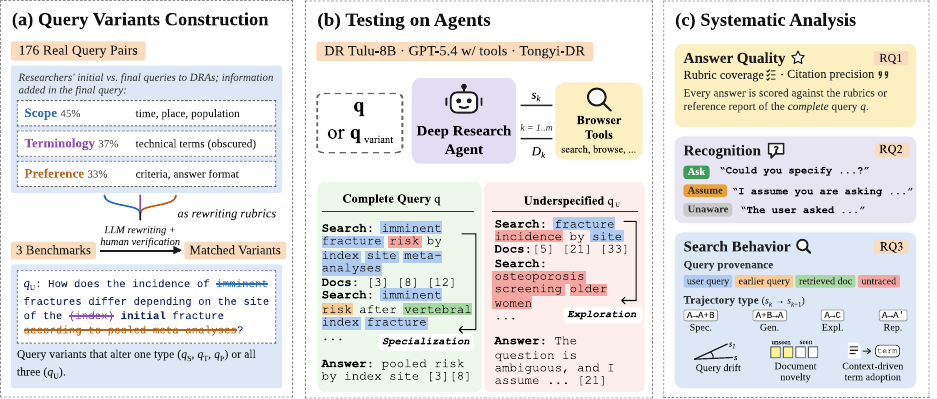}
\caption{Overview of \sys{}. (a) From 176 initial--final query pairs in researchers' conversations with DRAs, we identify three omitted information types: \textcolor{scopecolor}{\emph{scope}} ($q_s$), \textcolor{termcolor}{\emph{terminology}} ($q_t$), and \textcolor{prefcolor}{\emph{preference}} ($q_p$). We then rewrite benchmark queries into matched variants with an LLM followed by human verification. (b) Each agent answers the complete query $q$ or a variant $q_r$ by alternating search queries $s_k$ and retrieved documents $D_k$. Highlighted search-query tokens come from the user query (blue), earlier search queries (yellow), retrieved documents (green), or parametric knowledge (red). (c) We analyze answer quality (RQ1), whether agents recognize underspecification (RQ2), and search behavior
(RQ3).}
\vspace{-0.2in}
\label{fig:framework}
\end{figure*}

To this end, we introduce \sys{}, a framework for investigating DRA's answer quality and intermediate search behavior under user queries with controlled levels of underspecification, as illustrated in Figure~\ref{fig:framework}. To ensure that the underspecified queries reflect realistic forms of underspecification, we analyze initial--final query pairs collected from real-world researchers' conversation logs with DRAs. We identified three types of information that are frequently omitted: \textcolor{scopecolor}{\emph{scope constraints}} (e.g., temporal, spatial, or population constraints), \textcolor{termcolor}{\emph{terminology}} (e.g., technical or domain-specific terms), and \textcolor{prefcolor}{\emph{user preferences}} (e.g., evaluation criteria or the desired answer format). We then construct underspecified versions of three existing deep research benchmarks that features open-ended questions requiring long-form answers: \rqa{} \citep{li2025researchqa}, \sqa{} \citep{asai2024openscholar}, and \drb{} \citep{du2026deepresearchbench}.




We evaluate three widely used DRAs, DR Tulu \citep{shao2025drtulu}, Tongyi DeepResearch \citep{tongyi2025deepresearch}, and GPT 5.4 with the same tool set as DR Tulu, on both the complete queries and their underspecified variants. We assess how well their final answers satisfy the query-specific rubrics and how well their claims are supported by the cited evidence. Additionally, we examine whether DRAs recognize underspecified user query, what actions they take upon recognizing it, and whether these actions mitigate the impact of underspecification on final answer quality. To characterize changes in intermediate search behavior, we examine how search queries evolve across steps, whether successive searches introduce new documents, and whether retrieved documents inform subsequent searches. 

Our study contributes three main findings:
\begin{enumerate}[leftmargin=*,itemsep=0pt,topsep=0pt,parsep=1pt,label=(\arabic*)]
\item \emph{Query underspecification does not lead to more hallucination, but likely leads to answers that are misaligned with user's information needs.} Omitting or obscuring even one key detail like scope constraints or technical terms significantly reduces how well answers satisfy the query-specific rubrics. In contrast, even the most underspecified queries do not lead to more hallucinations.
\item \emph{DRAs fill in missing information based on their learned priors during training rather than explicitly eliciting the missing information from user.} They struggle to recognize when user queries lack critical information and, even when they do, they often proceed without seeking clarification from the user. This tendency is also reflected in their intermediate search behavior: their search queries rely less on the user’s original query and more on the DRA’s parametric knowledge.
\item \emph{Query underspecification leads to more diffuse search without greater use of retrieved evidence.} DRAs conduct broader searches with greater drift in their search query and retrieve more previously unseen documents. However, they do not make greater use of retrieved content in subsequent searches.



\end{enumerate}
 

 We hope our findings pave the way for developing DRAs that recognize unresolved information gaps and make them explicit before proceeding with the research task.
 


\section{Related Work}
\label{sec:related}

\paragraph{Deep research agents.} Deep research agents extend tool-augmented reasoning \citep{yao2023react} with iterative search and long-form synthesis, and are trained with reinforcement learning for search \citep{jin2025searchr1,li2025searcho1,li2025webthinker}, agentic mid- and post-training \citep{tongyi2025deepresearch}, or evolving rubrics \citep{shao2025drtulu}. Their benchmarks consist of fully specified queries \citep{mialon2024gaia,wei2025browsecomp,asai2024openscholar,li2025researchqa,du2026deepresearchbench}; we construct underspecified counterparts and analyze the research process in addition to the output.
\paragraph{Query underspecification and ambiguity resolution.} 
Prior work examines how omitted requirements affect LLM performance \citep{yang-etal-2026-prompts} and evaluates or improves agents’ ability to handle underspecified queries in software engineering \citep{vijayvargiya2026ambigswe}, question answering \citep{cao2026clarifyOA,su2026knowing,xie2026oversearching,du2026recognition}, and other tasks with easy-to-verify answers \citep{sun2026training,liu2026agentabstain}. In these settings, unresolved ambiguity can lead to observable task failures or deviations from user's expected outcomes, providing relatively clear signals of when clarification is needed. For open-ended deep research, however, report-style answers are difficult to verify, and the effects of query underspecification remain unclear. 

Given that human experts can select appropriate strategies to resolve ambiguity in such situations \citep{freeman2025zoom}, we seek to understand how to enable agents to develop similar capabilities. Prior work categorizes LLM's disambiguation policies into query rewriting, long-form answer generation, and asking clarifying questions \citep{tanjim2025disambiguation}, and highlights the need for agentic orchestration of different disambiguation strategies based on their trade-offs. Prior work explores uncertainty quantification to inform clarification decisions \citep{zhang2023clarify,kuhn2023semantic}, while other work uses preference optimization to align LLM responses with human experts’ communication practices, potentially improving question-asking behavior in specific domains \citep{seegmiller2026clinician}. However, extending these approaches to long-horizon agentic settings remains underexplored, where agents need to decide whether uncertainty can be resolved through further interaction with the environment or requires additional information from the user.

\section{The \sys{} Framework}
\label{sec:framework}

\begin{table*}[!ht]
\centering
\scriptsize
\setlength{\tabcolsep}{4pt}
\begin{tabular}{p{2.3cm} p{3.5cm} p{6.5cm} rrr}
\toprule
\multirow{2}{*}{Rubric} & \multirow{2}{3.5cm}{Removed / Obscured Information} & \multirow{2}{*}{Example} & \multicolumn{3}{c}{\shortstack{\# Query Variants}} \\
\cmidrule(lr){4-6}
& & & RQA & SQA & DRB \\
\midrule
\textcolor{scopecolor}{R1 Scope} ($\qS$) & Time, place, population, domain & Which regions and countries saw widespread acceptance and implementation of home delivery services for antiretroviral therapy (ART) \scopesout{during the COVID-19 pandemic}? & 274 & 129 & 74 \\
\textcolor{termcolor}{R2 Term} ($\qT$) & Precise technical terms (replaced by generic words) & How do non-aspirin \termsout{NSAIDs} \textbf{pain relievers} and selective \termsout{COX-2} \textbf{enzyme} inhibitors differ in their associations with aneurysmal subarachnoid hemorrhage risk, and what is known about the heterogeneity across studies for these drug classes? & 238 & 126 & 45 \\
\textcolor{prefcolor}{R3 Preference} ($\qP$) & Stated criteria, comparisons, or requirements on the answer & Which Obsidian plugins can effectively replicate Notion's multi-view database functionality (including Table, Kanban, Calendar, and List views)? \prefsout{Please provide a detailed comparison of the strengths and weaknesses of these plugins.} & 27 & 65 & 58 \\
Fully-Underspecified ($\qU$) & R1--R3 jointly & How does the incidence of \scopesout{imminent} fractures differ depending on the site of the \termsout{(index)} \textbf{initial} fracture \prefsout{according to pooled meta-analyses}? & 499 & 287 & 100 \\
\bottomrule
\end{tabular}
\caption{Underspecification rubrics, examples, and number of constructed variants in \rqa{} (RQA), \sqa{} (SQA), and \drb{} (DRB). Colors distinguish \textcolor{scopecolor}{scope}, \textcolor{termcolor}{terminology}, and \textcolor{prefcolor}{preference} edits; struck text is removed and bold text denotes replacements.}
\label{tab:rubrics}
\vspace{-10pt}
\end{table*}

\subsection{Problem Formulation}
\label{sec:formulation}
Let $q$ denote a fully specified query (complete query) that expresses a user's information need, and let $S_d(a;q)$ denote an evaluation function that scores answer $a$ on one dimension $d$ of answer quality (e.g., rubric coverage or citation precision). An underspecification operator $u_r$, with $r\in\{\textsc{s},\textsc{t},\textsc{p}\}$, removes \textcolor{scopecolor}{\emph{scope constraints}} ($r=\textsc{s}$) or obscures \textcolor{termcolor}{\emph{technical terminology}} ($r=\textsc{t}$) or \textcolor{prefcolor}{\emph{user preferences}} ($r=\textsc{p}$) in $q$, yielding the query variant $q_r=u_r(q)$. Each operator is applied only when the corresponding information is present in $q$ and the fully underspecified variant $\qU$ is obtained by applying all applicable operators jointly. Given a query, an agent $\pi$ produces a search trajectory $\tau=(s_1,D_1,\dots,s_m,D_m)$, where $s_k$ is the $k$-th search query and $D_k$ is the corresponding set of retrieved documents retrieved for it, follwed by a final answer $a_\pi(\cdot)$, respectively. We quantify the effect of underspecification on answer quality as
\begin{equation}
\label{eq:delta}
\Delta_{r,d}^{\mathrm{ans}}
=\mathbb{E}_{q}\big[S_d(a_\pi(q_r);q)-S_d(a_\pi(q);q)\big],
\end{equation}
where the expectation is taken over the queries to which $u_r$ applies.
By evaluating both answers against the complete query $q$, $\Delta_{r,d}^{\mathrm{ans}}$ measures how underspecification affects the agent's ability to satisfy the original information need.

Similarly, we quantify changes in intermediate search behavior as
\begin{equation}
\Delta_{r,l}^{\mathrm{search}}
=\mathbb{E}_{q}\big[M_l(\tau_\pi(q_r))-M_l(\tau_\pi(q))\big],
\end{equation}
where $M_l$ measures an aspect of the search behavior (Section~\ref{sec:setup}).
\subsection{Deriving Dimensions of Underspecification}
\label{sec:derive}
To ground the framework in real user behavior, we collect the conversation logs of three junior researchers in electrical engineering, computer science, and construction science fields with DR Tulu \citep{shao2025drtulu} over a two-week period. See details in Appendix~\ref{app:real-data}. We focused on junior researchers because they constitute a substantial segment of deep research agent users and may struggle to fully articulate their information needs due to limited domain expertise. From these logs, we select conversations in which they reformulated their initial queries after receiving an answer and we pair the initial query with the final query of each selected conversation and compare them to identify changes in wording and explicitly stated information. Of the 203 initial--final query pairs, 176 contained information in the final query that was absent from the initial query. 

We characterize the types of information omitted from users’ initial queries through inductive coding analysis \citep{thomas2006inductive}. The first and second authors independently examine each initial--final query pair, identify information that is newly specified in the final query, and assign descriptive codes to characterize the type of information added. They then compare their codes and categories, discuss discrepancies, and consolidate conceptually similar categories through consensus. 
This iterative process result in three higher-level categories of information that users commonly omitted from their initial queries: \textcolor{scopecolor}{\emph{scope}}, \textcolor{termcolor}{\emph{terminology}}, and \textcolor{prefcolor}{\emph{preferences}}. \textcolor{scopecolor}{\emph{Scope}} specifies the boundaries of the research question, such as the time period, place, population, or domain. \textcolor{termcolor}{\emph{Terminology}} identifies precise technical concepts or domain-specific entities. \textcolor{prefcolor}{\emph{Preferences}} specify the user's expectations for the answer, such as evaluation criteria, comparison requirements, or presentation format. Together, these three categories cover 94\% of the 176 pairs: missing \emph{scope} (45\%), missing \emph{terminology} (37\%), and missing \emph{preferences} (33\%). 21\% of the pairs lack more than one type of information.

 
\subsection{Constructing Matched Variants}

We formalize each category as a rewriting rubric (Table~\ref{tab:rubrics}) and prompt GPT-5.4 mini to generate underspecified query variants (see Appendix~\ref{app:query-rewrite-prompt} for the prompts). To verify whether these variants resemble naturally occurring underspecified queries, we conduct a Turing-style discrimination test \citep{turing1950computing}. We first apply the rewriting procedure to the clarified queries in our collected query pairs from real users. We then pair each generated variant with its corresponding human-written initial query and ask a separate model to identify which query is AI-generated and which is human-written.

We repeat the test twice, shuffling the order of the human-written and AI-generated queries in each run. Across 486 generated variants derived from 176 clarified queries, GPT-4o achieves identification accuracies of 58\% and 54\%, respectively, slightly higher than random guessing. This result suggests that the generated variants are reasonably similar to naturally occurring underspecified queries, and that the rewriting procedure can be used to construct matched query variants for existing deep research benchmarks.

We apply this rewriting procedure to three deep research benchmarks: 500 randomly sampled queries from \rqa{}, the ScholarQA-CS, ScholarQA-Multi, and ScholarQA-CS2 subsets of \sqa{}, and all queries of \drb{}. Table~\ref{tab:rubrics} reports the number of resulting variants for each rubric.

 
\subsection{Human Validation of Synthetic Query Variants}

The first and second authors each validated half of the generated query variants, checking whether the targeted information was removed or obscured, non-target information was preserved, and the resulting query remained natural and coherent. Overall, 6\% of the variants were revised to improve naturalness, and 8\% were revised to ensure that technical terminology was appropriately obscured.

\section{Experimental Setup}
\label{sec:setup}

\paragraph{Models and harnesses.} We evaluate three common deep research agents: \textbf{\tulu{}} \citep{shao2025drtulu}, which operates within an MCP-based agent framework with three tools: web search, scientific paper search, and webpage browsing. Its interaction protocol supports
reasoning, tool calling, answer generation, and citation. The second one is \textbf{\tulugpt{}}, which uses the same agent framework and tool set as DR Tulu. \textbf{\tongyi{}} \citep{tongyi2025deepresearch}
is equipped with tools for web search, webpage
browsing, Google Scholar search, Python execution,
and file parsing. Unlike \tulu{}, which submits a single query per search tool call, \tongyi{} supports multiple queries within a single search tool call.
 
\begin{figure*}[!tb]
\centering
\includegraphics[width=\textwidth]{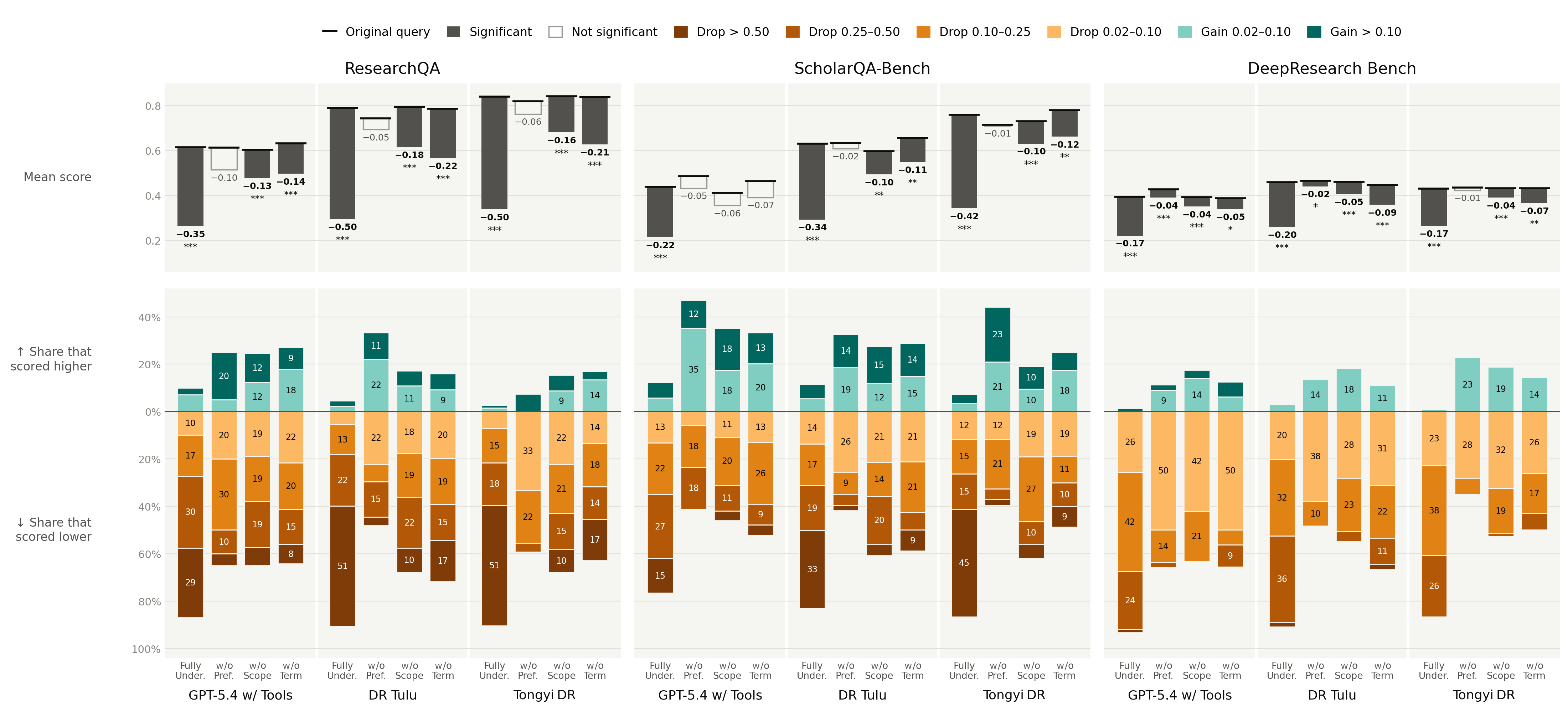}
\caption{Effects of query underspecification on answer quality across three benchmarks and DRAs. \textbf{Top:} Mean paired score changes (underspecified variant minus complete query). Filled and outlined bars indicate significant and nonsignificant differences, respectively, based on paired Wilcoxon signed-rank tests ($^{*}p<.05$, $^{**}p<.01$, $^{***}p<.001$). \textbf{Bottom:} Proportions of matched queries with lower scores (below zero) or higher scores (above zero), with colors indicating the magnitude of change. Segment labels are percentages; score changes within $\pm0.02$ are omitted.}
\label{fig:quality}
\label{fig:coverage}
\end{figure*}

\paragraph{Benchmarks and answer quality metrics.} We use three benchmarks of open-ended research questions requiring long-form answers: 500 randomly sampled queries from \rqa{} \citep{li2025researchqa}; the ScholarQA-CS, ScholarQA-Multi, and ScholarQA-CS2 subsets of \sqa{} \citep{asai2024openscholar}; and \drb{} \citep{du2026deepresearchbench}. We use Gemini 3.1 Pro as the judge to score each answer for \textbf{\emph{rubric-based coverage}}, following each benchmark's official evaluation protocol: query-specific rubrics for \rqa{} and \sqa{}, and general rubrics with a reference answer for \drb{}. These scores measure how well an answer fulfills the original query's information needs. We also assess the extent to which answers are grounded in retrieved evidence, using \textbf{\emph{Document utilization}} to measure how much of the retrieved material is cited and \textbf{\emph{Citation precision}} to measure whether the cited evidence supports the answer's claims. For each answer, we compute
$\text{Document Utilization} = \#_{\mathrm{cited}} / \#_{\mathrm{retrieved}}$ and $\text{Citation Precision} = \#_{\mathrm{supporting}} / \#_{\mathrm{cited}}$ where $\#_{\mathrm{retrieved}}$ is the total number of documents retrieved during the whole search process, $\#_{\mathrm{cited}}$ is the number of documents cited in the final answer, and $\#_{\mathrm{supporting}}$ is the number of cited documents that support the claims for which they are cited. To compute $\#_{\mathrm{supporting}}$, we use Gemini 3.1 pro as a judge to examine the content of each cited document and assess whether it supports the associated claims in the answer. Only \sqa{} and \drb{} support separate evaluation of citation quality, so we evaluate citation-related metrics only for \tulu{} and \tulugpt{} on these two benchmarks, excluding \tongyi{} because it does not include explicit citations in its final answers. For rubric-based coverage, we use Wilcoxon signed-rank tests with Bonferroni correction for 12 comparisons per agent (three benchmarks $\times$ four variants) to assess evidence of systematic paired score changes. For document utilization and citation precision, we report unadjusted Wilcoxon signed-rank tests, which is conservative for our conclusion that these metrics do not decline.

\begin{table*}[!ht]
\centering
\scriptsize
\setlength{\tabcolsep}{4pt}
\begin{tabular*}{\textwidth}{@{\extracolsep{\fill}}llcccc@{}}
\toprule
Model & Metric & Fully underspec. & w/o preference & w/o scope & w/o term \\
\midrule
\multirow{2}{*}{\texttt{DR Tulu}} & Document Utilization & $+2.85\%^{***}$ & $+0.01\%$ & $+1.51\%^{*}$ & $+0.07\%$ \\
 & Citation Precision & $+0.45\%$ & $+1.40\%$ & $-2.88\%$ & $+2.17\%$ \\
\midrule
\multirow{2}{*}{\texttt{GPT-5.4 w/ Tools}} & Document Utilization & $-0.44\%$ & $+1.98\%$ & $+2.11\%^{**}$ & $-0.45\%$ \\
 & Citation Precision & $-2.14\%$ & $+4.68\%$ & $-0.48\%$ & $+0.83\%$ \\
\bottomrule
\end{tabular*}
\caption{Mean paired changes in document utilization in the final answer and citation precision (underspecified variant minus complete query), pooled across datasets within each model. Values are displayed as percentages and represent percentage-point differences. Significance is assessed using unadjusted paired Wilcoxon signed-rank tests. Stars denote $^{*}p<0.05$, $^{**}p<0.01$, and $^{***}p<0.001$.}
\label{tab:citation-precision-diff}
\end{table*}

\paragraph{Search behavior metrics.}
\label{sec:measures}
We adopt measures from prior work on agentic search behavior analysis. \textbf{\emph{Query drift}} \citep{ning2026agentic} measures the semantic shift between
the first and last search queries in the whole deep research process. For a trajectory containing
$m$ search queries, we compute drift as
$1-\cos(\mathbf{v}_1,\mathbf{v}_m)$, where
$\mathbf{v}_1$ and $\mathbf{v}_m$ are the embeddings
of the first and final search queries, respectively,
obtained using OpenAI's \texttt{text-embedding-3-small} model. Higher values indicate greater semantic drift. \textbf{\emph{Query trajectory type}} \citep{ning2026agentic}
characterizes how consecutive search queries are reformulated.
We use GPT-5.4 mini to classify each transition $s_k \rightarrow s_{k+1}$ into one of four types: \emph{Specialization} (adding constraints
to narrow the search), \emph{Generalization} (relaxing
constraints to broaden the search), \emph{Exploration}
(shifting to a different facet of the same topic), or
\emph{Repetition} (issuing an identical or near-duplicate
query). We report the proportion of transitions of each type in a complete trajectory. \textbf{\emph{Query provenance}} \citep{zhang2026hawkeye} attributes each token of $s_k$ to the first of four sources in which it occurs: the \emph{user query}, the \emph{previous search queries} $s_{<k}$, the \emph{retrieved documents} $D_{<k}$, or none of these (\emph{untraced}; attributed to parametric knowledge or memory). Provenance shares are computed per search tool call and then averaged. \textbf{\emph{Document novelty}} \citep{zhang2026hawkeye}
measures the fraction of documents retrieved at each
search step that have not appeared in earlier steps.
Higher values indicate more newly retrieved documents,
whereas lower values indicate greater retrieval redundancy.
\textbf{\emph{Context-driven term adoption rate}}
\citep{ning2026agentic} measures the fraction of newly introduced search-query terms that appear in previously retrieved documents. Higher values indicate greater lexical adoption of retrieved evidence in subsequent searches.
For search behavior measures, we use paired $t$-tests on per-trajectory values; Wilcoxon signed-rank tests yield the same conclusions. We report Cohen's $d_z$ as the effect size \citep{cohen1988statistical}.

\section{RQ1: How Does Underspecification Affect Answer Quality?}
\label{sec:coverage}

{
\textbf{DRAs cannot reliably infer users' information needs.}\enspace
Figure~\ref{fig:quality} shows that rubric-based coverage scores decline for a large proportion of underspecified queries even when only one key detail is omitted or obscured. Among different information types, removing scope constraints or obscuring technical terminology produces substantial mean score reductions that are statistically significant across most agents and benchmarks. In contrast, removing user preferences generally produces smaller reductions that are often not statistically significant. As user queries become more underspecified, scores decline further.
\par}
{
\textbf{Query underspecification does not lead to more hallucination.}\enspace Table~\ref{tab:citation-precision-diff} shows no statistically significant change in citation precision even under full underspecification condition. Document utilization in the final answer also shows no statistically significant decline: it even increases in three of the eight settings. Together with the decline in rubric-based coverage, these results suggest that underspecification primarily reduces alignment with the user's original information needs, with no evidence of increased hallucination among cited claims.
\par}

{
\textbf{Stronger specifically-trained deep research capability come with greater sensitivity to query underspecification.}\enspace Using performance on complete queries as a measure of deep research capability, Figure~\ref{fig:quality} shows that higher-scoring agents generally incur larger absolute score losses under full underspecification. Both \tulu{} (8B) and \tongyi{} (30B) are relatively compact models trained specifically for deep research. They outperform GPT-5.4 with tools on complete queries but appear to depend more heavily on users fully specifying their information needs.
\par}

\section{RQ2: Can DRAs Recognize Underspecification in User Queries?}
\label{sec:ask}
\begin{figure*}[!tb]
\centering
\begin{minipage}[t]{0.39\textwidth}
    \centering
    \includegraphics[width=\linewidth]{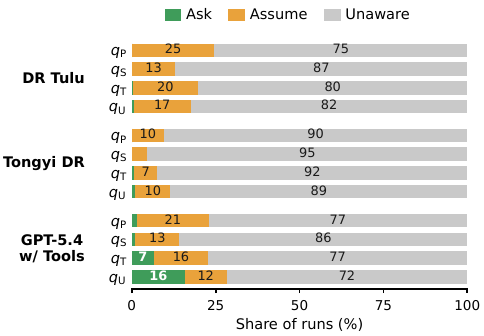}
    \caption{Clarification behavior under underspecification across three benchmarks. Agents either ask for clarification (Ask), state assumptions (Assume), or proceed without recognizing omissions (Unaware).}
    \label{fig:clarify}
\end{minipage}
\hfill
\begin{minipage}[t]{0.59\textwidth}
    \centering
    \includegraphics[width=\linewidth]{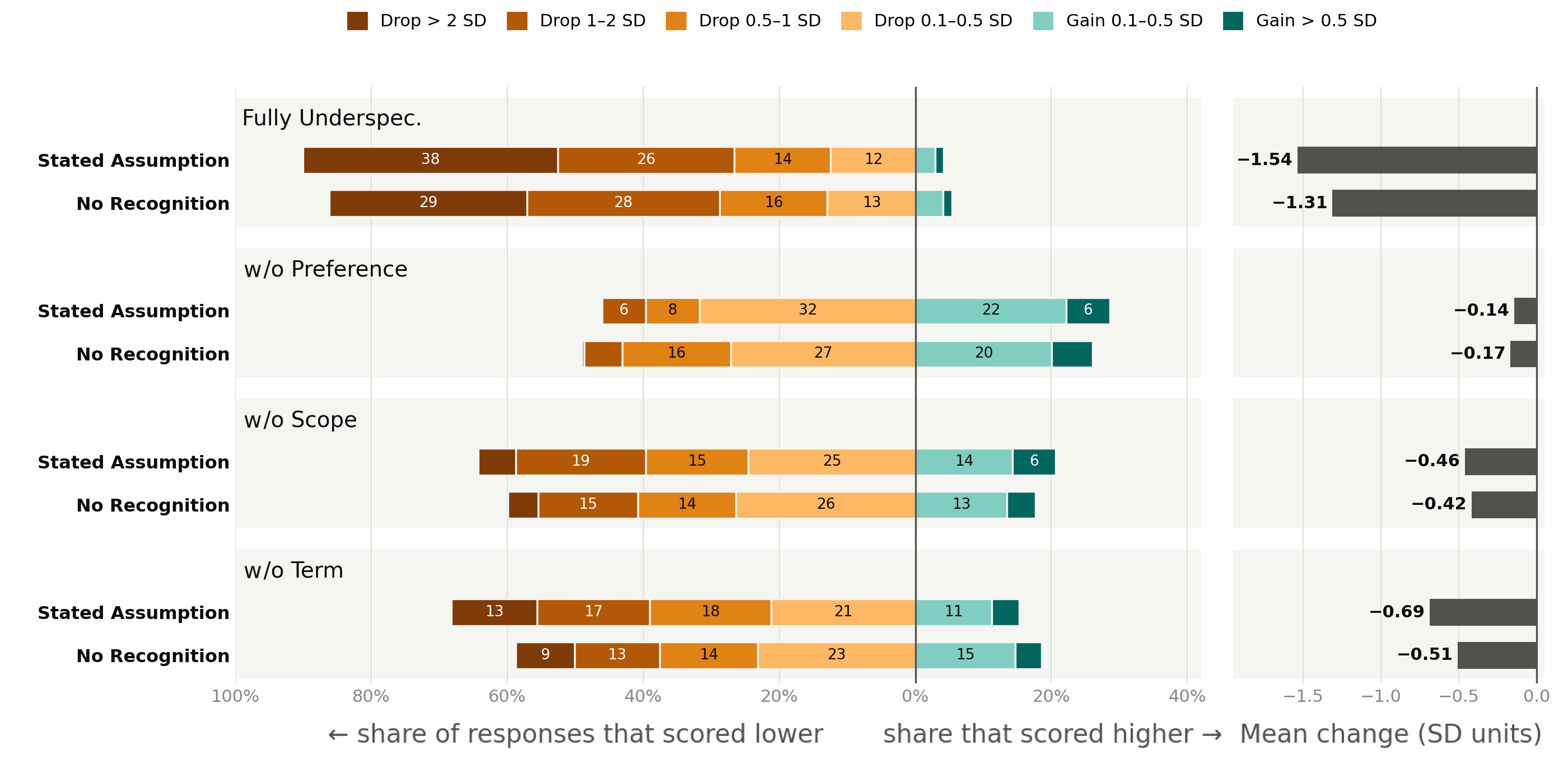}
    \caption{Scaled performance drop by underspecification type, pooled across all three benchmarks and models, comparing runs that state an assumption and proceed with runs that do not recognize the ambiguity.}
    \label{fig:ambiguity-drop-by-ablation-type}
\end{minipage}
\vspace{-0.1in}
\end{figure*}
\label{sec:query}
We observed that DRAs sometimes abstain from answering underspecified queries, even though we did not indicate that the user query may be underspecified in system prompt. We therefore analyze these behaviors and identified three mutually exclusive categories: \textbf{Unaware}, in which the DRA proceeds to answer without explicitly recognizing the underspecification; \textbf{Assume}, in which it recognizes the underspecification but proceeds by making one or more explicit assumptions; and \textbf{Ask}, in which it abstains from answering and instead asks the user for clarification. See detailed analysis protocol in Appendix \ref{app:analysis}. We then use GPT-5.4 mini to classify all the responses into these categories based on observable behavior in the DRA’s reasoning trajectories and final answers.

\textbf{DRAs show limited awareness of missing information.}\enspace Although answer quality declines for a large proportion of underspecified queries  (Figure \ref{fig:coverage}), DRAs generally cannot identify that key information is missing: Unaware is the most frequent category across all different DRAs (Figure~\ref{fig:clarify}; Table~\ref{tab:clarify-full}). Even when they acknowledge ambiguity, they usually answer under one or more stated assumptions rather than ask the user for clarification. Across agents, \tongyi{} shows the least explicit awareness, whereas \tulu{} and \tulugpt{} more often state assumptions.

\textbf{Stated assumptions do not guarantee better answers, and even worse.}\enspace  Although stating assumptions is DRAs' predominant response strategy upon recognizing underspecification, Assume runs exhibit larger answer-quality losses than Unaware runs in three of the four conditions, as shown in Figure \ref{fig:ambiguity-drop-by-ablation-type}. One possible explanation is that unrecognized underspecification may be mild enough for DRAs to infer a plausible interpretation, whereas explicitly recognized ambiguity may reflect greater uncertainty about the user's intent. In such cases, proceeding under assumptions may be less appropriate than seeking clarification.



\section{RQ3: How Does Underspecification Affect Intermediate Search Behavior?}
\begin{figure*}[!tb]
\centering
\begin{minipage}[t]{0.49\textwidth}
    \centering
    \includegraphics[width=\linewidth]{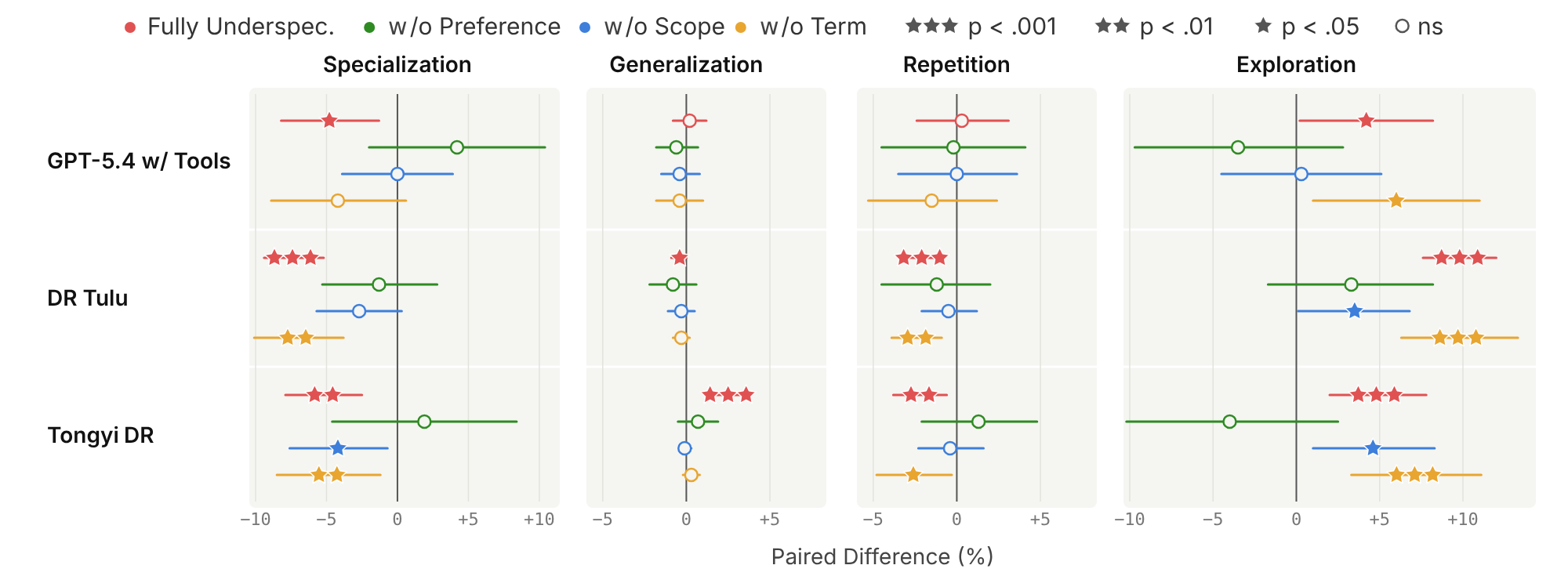}
    \caption{Changes in search-query transition types under underspecification. Values show mean paired differences (underspecified minus complete query, in percentage points) with 95\% confidence intervals. Positive values indicate more frequent transitions. Stars denote significance (* $p<.05$, ** $p<.01$, *** $p<.001$).}
    \label{fig:search}
\end{minipage}
\hfill
\begin{minipage}[t]{0.49\textwidth}
    \centering
    \includegraphics[width=\linewidth]{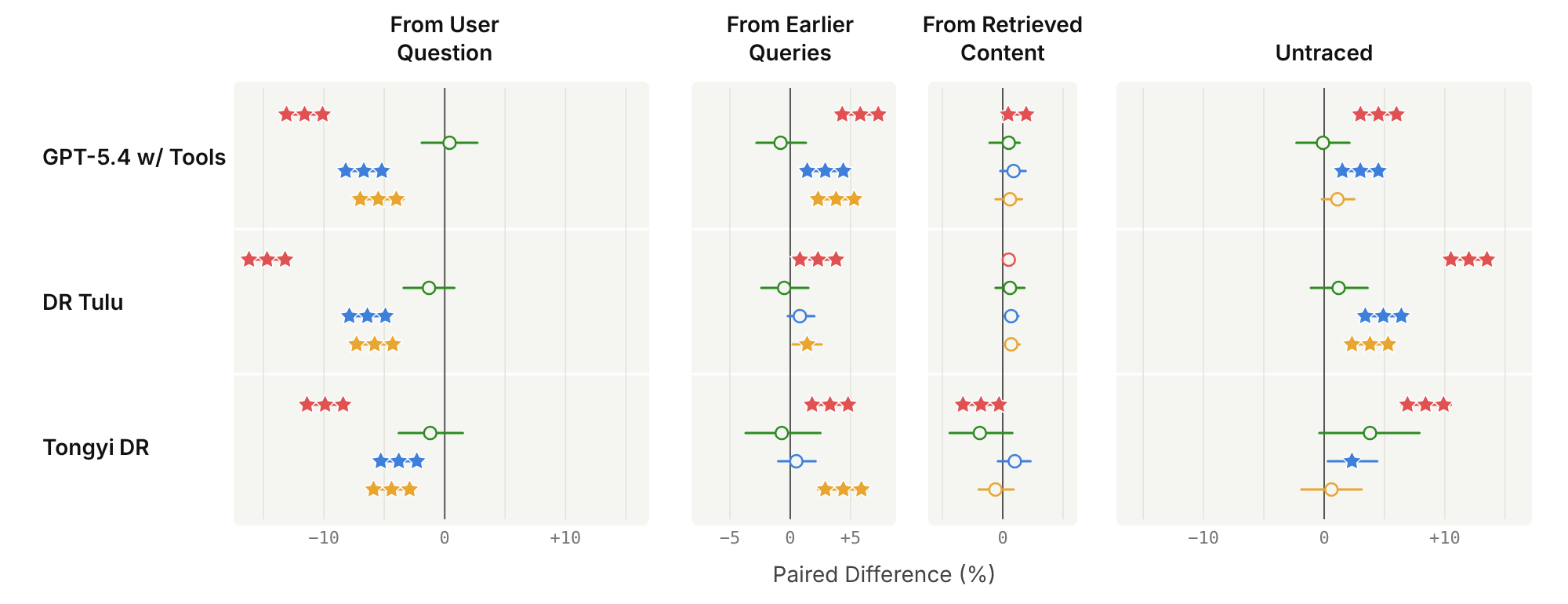}
    \caption{Changes in search-query token provenance under underspecification. Values show mean paired differences in token shares (underspecified minus complete query, in percentage points) with 95\% confidence intervals. Positive values indicate larger shares. Significance markers follow Fig.~\ref{fig:search}.}
    \label{fig:query-provenance}
\end{minipage}
\end{figure*}

\begin{figure}[t]
\centering
\includegraphics[width=\columnwidth]{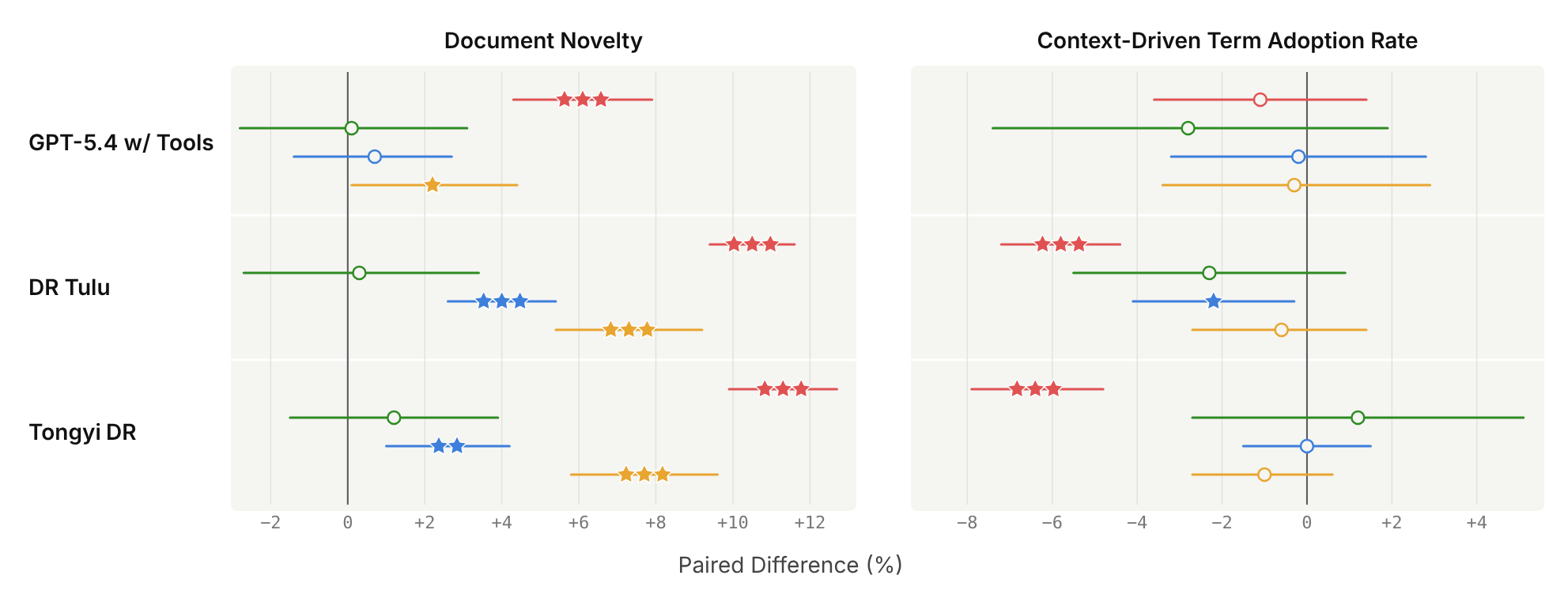}
\caption{Changes in document novelty and context-driven term adoption under underspecification. Values show mean paired differences with 95\% confidence intervals. Positive novelty changes indicate more newly retrieved documents; negative adoption changes indicate less use of retrieved content. Significance markers follow Fig.~\ref{fig:search}.}
\label{fig:novelty-adoption}
\vspace{-10pt}
\end{figure}

\textbf{DRAs change search direction more often and investigate less deeply.}\enspace This shift is most evident under the most underspecified variant, with more Exploration and fewer Specialization transitions across all three DRAs (Figure~\ref{fig:search}). Obscuring technical terminology also has a pronounced effect on search strategy. These changes indicate more frequent switches between different facets of a topic and fewer refinements that add constraints to narrow the search. Changes in the proportions of Generalization and Repetition are generally smaller and are not statistically significant in most comparisons.


\textbf{Searches become less grounded in user queries and draw more on parametric knowledge.}\enspace According to Figure \ref{fig:query-provenance}, the share of search-query tokens traceable to the user query decreases significantly, while the share of untraced tokens increases significantly in most cases. These untraced tokens occur in neither the user query, earlier search queries, nor retrieved documents, suggesting potential greater reliance on parametric knowledge to formulate searches. Scope removal produces a more consistent increase in the untraced token share than obscuring terminology. Removing preferences does not significantly change token provenance of search queries.

\textbf{Broader retrieval does not translate into greater utilization of retrieved content in subsequent searches.}\enspace Figure~\ref{fig:novelty-adoption} has shown that document novelty increases significantly under fully underspecified conditions for all three DRAs and without term, without scope conditions for DR Tulu and Tongyi. However, the context-driven term adoption rate does not increase correspondingly, and even dropped significantly for DR Tulu and Tongyi under the most underspecified conditions.

\section{Discussion and Conclusion}
\label{sec:discussion}
 
\paragraph{Users should prioritize clearly defining the research question when formulating search queries.} In particular, users should state key scope constraints and use precise technical terminology, as omitting or obscuring this information can substantially affect the search process and degrade the final answer quality. By comparison, information like user preferences has a smaller effect on answer quality.

 \paragraph{DRAs need to be explicitly trained to identify missing information during long-horizon task execution.} Our experiments revealed the usability challenge of current DRAs: users may need sufficient domain knowledge to identify relevant constraints and terminology before agents can effectively address their needs. For lay person who discovers missing specifications only after reviewing a lengthy report, they must spend additional time diagnosing the mismatch and reformulating their requests. Repeating the research process also incurs potentially avoidable tool calls and token consumption. We argue that DRAs should distinguish uncertainty that can be resolved through further search from missing information that must be elicited from the user. Training this capability through imitation learning requires training data that capture both deep research trajectories and appropriate clarification actions, but such data can be difficult to obtain. Our analysis of intermediate search behavior suggests that metrics such as the proportion of untraced search-query tokens, and context-driven term adoption rate may serve as useful indicators of missing information in user queries.

 We introduce \sys{}, a framework for evaluating Deep Research Agents under query underspecification. Using this testbed, we examine changes in final answer quality and intermediate search behavior casued by query underspecification. We further analyze DRAs' awareness of query underspecification, their response strategies upon recognizing it, and whether these strategies mitigate answer-quality losses. We find that evidence-grounded reports may still fail to meet users' needs.These results provide practical advice on how users should prompt DRAs to get desired answer and offer implications for building DRAs that recognize when to ask users for additional input, balancing reliable task completion with minimal disruption.

\FloatBarrier

\section*{Limitations}
\paragraph{Language.} Our experiments primarily use English queries; DeepResearch Bench includes tasks in Chinese. Underspecification may take different forms in other languages and extending \sys{} to these settings is a natural direction for future work.

\paragraph{Coverage of agents.} Our evaluation covers three agents whose intermediate search and tool-use trajectories are accessible for analysis. We do not evaluate closed-source deep research systems that do not expose these trajectories, as this would prevent us from analyzing their search behavior. These systems may employ different search and clarification strategies.

\section*{Ethics Statement}
Our study does not involve the generation of harmful content. Its main implication is a risk for users: when a query is underspecified, current DRAs may return a well-cited answer to a broader or different question than the one the user intended, usually without signaling that information was missing. Users who rely on such answers, for example in medical or policy research, may draw conclusions that do not apply to their situation. We hope that our findings encourage the development of agents that recognize missing information and make their assumptions visible to users.



\bibliography{custom}

\appendix
\makeatletter
\setlength{\@dblfptop}{0pt}
\setlength{\@dblfpsep}{12pt}
\setlength{\@dblfpbot}{0pt plus 1fil}
\makeatother
\section{Prompt Templates}
\label{app:examples}

\subsection{Constructing Underspecified Query Variants}
\label{app:query-rewrite-prompt}
The rewriting procedure first checks whether a query contains preference information, terminology, or scope constraints, then generates the applicable underspecified variants. We reproduce the default prompts from the script below. The placeholder \texttt{[QUERY]} denotes the original query.


\noindent\textbf{Information detection: system prompt.}\par
\begin{nolinenumbers}
\begin{lstlisting}[style=prompt]
You analyze research queries.

Return exactly one JSON object and nothing else.
The JSON object must have these keys:
- "has_preference"
- "has_term"
- "has_scope_limitation"

For each key, the value must be exactly "yes" or "no".
\end{lstlisting}
\end{nolinenumbers}

\noindent\textbf{Information detection: user prompt.}\par
\begin{nolinenumbers}
\begin{lstlisting}[style=prompt]
Analyze the query and answer these three questions:

1. "has_preference": Does the query explicitly contain user preference information, such as desired format, depth, target audience, or evaluation criteria?
2. "has_term": Does the query contain any proper nouns, such as named systems, algorithms, frameworks, organizations, or specific named entities?
3. "has_scope_limitation": Does the query explicitly restrict the problem to a specific scenario, domain, time period, or location? 


Query:
[QUERY]
\end{lstlisting}
\end{nolinenumbers}

\noindent\textbf{Query rewriting: shared system prompt.}\par
\begin{nolinenumbers}
\begin{lstlisting}[style=prompt]
You need to rewrite user queries.

You will be given:
1. The original query
2. One category of information to simplify

Return only the rewritten query. Do not explain the changes.
\end{lstlisting}
\end{nolinenumbers}

\noindent\textbf{Preference removal: user prompt.}\par
\begin{nolinenumbers}
\begin{lstlisting}[style=prompt]
Remove any user preference information in the original query(e.g. format, depth, audience, criteria).

Query:
[QUERY]
\end{lstlisting}
\end{nolinenumbers}

\noindent\textbf{Terminology simplification: user prompt.}\par
\begin{nolinenumbers}
\begin{lstlisting}[style=prompt]
Simplify any proper nouns (e.g. named technology categories, algorithms, frameworks, organizations) to a brief generic mention.

Query:
[QUERY]
\end{lstlisting}
\end{nolinenumbers}

\noindent\textbf{Scope removal: user prompt.}\par
\begin{nolinenumbers}
\begin{lstlisting}[style=prompt]
Remove any explicit scope limitations (e.g. scenario, domain, time period, location).

Query:
[QUERY]
\end{lstlisting}
\end{nolinenumbers}

\noindent\textbf{Full underspecification: user prompt.}\par
\begin{nolinenumbers}
\begin{lstlisting}[style=prompt]
Remove all the important details in the user query, such as user preference information (e.g. format, depth, audience, criteria), proper nouns (e.g. named technology categories, algorithms, frameworks, organizations), and scope limitations (e.g. scenario, domain, time period, location). Replace named entities with generic descriptions where needed. 

Query:
[QUERY]
\end{lstlisting}
\end{nolinenumbers}

\subsection{Underspecification Awareness and Response Classification}
\label{app:query-ambiguity-prompt}
The following prompt classifies whether an agent recognizes query underspecification and how it responds, assigning one of three categories: Ask, Assume, or Unaware. The placeholders \texttt{[QUERY]} and \texttt{[AGENT\_RESPONSE]} represent the query received by the agent and its full response, respectively.

\begin{nolinenumbers}
\begin{lstlisting}[style=prompt]
You are evaluating how a deep research AI agent handled a user's query.

Below is the ORIGINAL USER QUERY the agent was given, followed by the AGENT'S FULL RESPONSE.

Your task: determine whether the agent recognized that the user's query was ambiguous, underspecified, or missing information needed to answer it well, AND how it reacted to that.

Classify the agent's behavior into exactly ONE of these three categories:
- "asked_clarifying_question": the agent explicitly asked the user one or more questions to clarify the request, before or instead of giving a full answer.
- "stated_assumption_and_proceeded": the agent recognized the ambiguity, proposed one or more explicit assumptions/interpretations to resolve it, and then went ahead and produced a full (typically long) answer on that basis.
- "no_recognition": the agent showed no sign of recognizing any ambiguity in the query and just answered as if the query were fully specified.

Also provide:
- "evidence": a short (at most 2 sentences) quote or close paraphrase from the response that supports your classification, in the original language of the response. If the category is "no_recognition", instead give a one-sentence explanation of why there is no sign of recognition.

Respond with ONLY a single JSON object, no other text, no markdown code fences:
{"category": "asked_clarifying_question" | "stated_assumption_and_proceeded" | "no_recognition", "evidence": "..."}

<original_user_query>
[QUERY]
</original_user_query>

<agent_response>
[AGENT_RESPONSE]
</agent_response>
\end{lstlisting}
\end{nolinenumbers}

\subsection{Search-Query Transition Classification}
\label{app:query-transition-prompt}
The following prompts classify the transition between two consecutive search queries. The placeholders \texttt{\{q1\}} and \texttt{\{q2\}} are replaced with the previous and current search queries, respectively.

\noindent\textbf{System prompt.}\par
\begin{nolinenumbers}
\begin{lstlisting}[style=prompt]
You are an expert search behavior analyst.
\end{lstlisting}
\end{nolinenumbers}

\noindent\textbf{User prompt.}\par
\begin{nolinenumbers}
\begin{lstlisting}[style=prompt]
Query 1 (Previous): {q1}
Query 2 (Current): {q2}

Analyze the search behavior evolution from Query 1 to Query 2 for an autonomous agent.
Classify the transition into exactly ONE of these four categories:

1. Specialization (Vertical Deepening): Query 2 is MORE specific than Query 1 by adding constraints/details (q2 \subset q1). (e.g., 'apple' -> 'green apple nutritional value').
2. Generalization (Vertical Broadening): Query 2 is MORE general than Query 1 by removing constraints/abstracting (q2 \supset q1). (e.g., 'green apple nutritional value' -> 'benefits of fruits').
3. Exploration (Horizontal Expansion within the same domain/task): Query 2 is NOT simply more specific or more general. It shifts to a different aspect / subtopic / related entity but still within the same overall topic/domain. (e.g., 'green apple nutritional value' -> 'green apple recipes', or 'MRI Scans' -> 'CT Scans').
4. Repetition (Stationary): Query 2 is semantically equivalent to Query 1. It is a paraphrase, reformatting, or synonym replacement with NO significant change in intent (e.g., 'green apple value' -> 'nutritional value of green apple').

Output ONLY the category name (Specialization, Generalization, Exploration, or Repetition).
\end{lstlisting}
\end{nolinenumbers}

\section{Deep Research Agent Configuration}
\label{app:agent-config}

We run all three deep research agents with the same budget of at most 15 tool calls per query. All inference is run on four NVIDIA H100 GPUs. The agent-specific configurations are as follows:
\begin{itemize}[leftmargin=*,itemsep=2pt,topsep=2pt]
\item \textbf{DR Tulu}: uses DR-Tulu-8B as the backbone model. Web search is performed through the Serper API and scholarly search through the Semantic Scholar API, and web pages are browsed through the Jina API.
\item \textbf{\tulugpt{}}: uses the same agent framework and tool set as \tulu{}, with GPT-5.4 as the backbone model. Search, scholarly search, and browsing therefore use the Serper, Semantic Scholar, and Jina APIs, respectively.
\item \textbf{\tongyi{}}: uses Tongyi-DR-30B-A3B as the backbone model. Web search is performed through the Serper API, and web pages are browsed through the Jina API.
\end{itemize}

\section{Real-World Query Collection Protocol}
\label{app:real-data}

This section describes the protocol used to collect and analyze real user queries in Section~\ref{sec:derive}.

Participants used the public web demo of DR Tulu released by AI2 (\url{https://www.dr-tulu.org/chat}). We asked them to ask agent questions arising from their own research. Then they were asked to carefully check whether each response matched their expectations and, if it did not, revise the query and ask again, repeating until they obtained a satisfactory response. Because the deep research agent does not support multi-turn interaction, each reformulation was submitted as a new, self-contained query rather than as a follow-up message. For every question that was not answered satisfactorily on the first attempt, participants shared the corresponding chat history via a link. All participants gave their consent to the use of their shared chat histories in this study. 

\section{Answer Analysis Protocol}
\label{app:analysis}

We first manually inspected approximately 10\% of the 1,922 agent responses to underspecified queries (about 192 responses). We sampled proportionally within each query-variant type, selecting approximately 10\% of responses from each of the fully underspecified, without scope, without terminology, and without preferences conditions. This inspection revealed that some agents acknowledged ambiguity but continued answering under explicit assumptions. We therefore adapt the behavioral analysis approach of \citet{su2026knowing} to distinguish three mutually exclusive categories, based on each agent's input query, recorded reasoning trajectory, and final response: \textbf{Ask},  \textbf{Assume}, \textbf{Unaware}.

We then prompted GPT-5.4 mini to annotate the same 192 responses using these categories. Cohen's $\kappa$ between the human and model annotations is 0.92, indicating almost perfect agreement. Following this validation step, we used the same model and prompt to annotate all responses. The classification prompt is provided in Appendix~\ref{app:query-ambiguity-prompt}, and each label is accompanied by a brief justification.

\section{Examples of Underspecified Queries}
\label{app:query-examples}

We contrast four examples from \rqa{} in which removing scope information or obscuring technical terminology has different effects on the answers generated by \tulu{}. Both the complete-query answer and the underspecified-query answer are evaluated against the original query's rubrics by Gemini 3.1 Pro. Struck-through text marks removed scope information in the without-scope variants ($\qS$). For the without-terminology variants ($\qT$), struck-through text marks the original technical terms and bold text gives their replacements.

\subsection{Lower Rubric Score After Removing Scope from the User Query}
\label{app:example-lower-coverage}

\begin{quote}
What are the main clinical advantages and safety considerations of using nebulized tranexamic acid to manage \scopesout{post-tonsillectomy} bleeding?
\end{quote}

\paragraph{Complete-query answer (score: 0.8333).}
The answer focuses on post-tonsillectomy hemorrhage. It discusses local bleeding control, reduced need for reoperation and blood products, pediatric care pathways, and escalation to surgery for persistent bleeding. Its safety discussion also addresses monitoring and precautions in this clinical setting.

\paragraph{Without-scope answer (score: 0.4167).}
After the surgical context is removed, the answer shifts toward hemoptysis and pulmonary hemorrhage, with additional discussion of epistaxis and sinonasal procedures. For example, it emphasizes ``higher 30-minute cessation rates than IV TXA'' in emergency hemoptysis. Although it still discusses bleeding control and safety, its focus moves away from the original post-tonsillectomy setting. This shift accompanies a decrease in rubric coverage from 0.8333 to 0.4167.

\subsection{No Decrease in Rubric Score After Removing Scope from the User Query}
\label{app:example-unchanged-coverage}

\begin{quote}
How does methotrexate work to modulate immune and inflammatory pathways in the treatment of \scopesout{moderate-to-severe} atopic dermatitis?
\end{quote}

\paragraph{Complete-query answer (score: 1.0000).}
The answer describes methotrexate's effects on folate metabolism, immune-cell proliferation, inflammatory cytokines, and epidermal biology, explicitly discussing moderate-to-severe atopic dermatitis.

\paragraph{Without-scope answer (score: 1.0000).}
The answer continues to discuss these mechanisms and retains the omitted severity context. For example, its clinical-evidence discussion states that treatment is associated with ``clinical improvement in moderate--severe AD.'' It also discusses severe or refractory disease. Thus, removing the severity qualifier does not prevent the answer from covering the original query's rubric requirements, and the score remains 1.0000. This example illustrates that omission of scope information does not necessarily cause a loss in rubric coverage when the answer still addresses the intended context.

\subsection{Lower Rubric Score After Obscuring Terminology in the User Query}
\label{app:example-term-lower-coverage}

\begin{quote}
How do different temperatures affect the formation and longevity of trapping nets in \termsout{Duddingtonia flagrans} \textbf{a fungus}?
\end{quote}

\paragraph{Complete-query answer (score: 0.7500).}
The answer focuses on \emph{Duddingtonia flagrans}, discussing temperature-dependent trap formation, signaling, structural persistence, and the use of spores in biological control. It organizes its discussion around the formation and longevity of trapping nets in the named species.

\paragraph{Without-terminology answer (score: 0.2500).}
Replacing the species name with ``a fungus'' leads the answer to consider multiple interpretations. It discusses nematode-trapping fungi such as \emph{Arthrobotrys} and also introduces a discussion of structures in basidiomycete fruiting bodies. The answer explicitly frames its response as interpreting trapping nets in ``two closely related, evidence-rich systems.'' Its focus thus broadens beyond the species specified in the original query, and rubric coverage falls from 0.7500 to 0.2500.

\subsection{No Decrease in Rubric Score After Obscuring Terminology in the User Query}
\label{app:example-term-higher-coverage}

\begin{quote}
How does chemical activation with \termsout{sodium hydroxide} \textbf{a base} improve the adsorption capacity of \termsout{zeolite} \textbf{a mineral adsorbent} for biogas upgrading?
\end{quote}

\paragraph{Complete-query answer (score: 0.59375).}
The answer discusses sodium-hydroxide activation of zeolites, emphasizing surface chemistry, ion exchange, and the adsorption of hydrogen sulfide and carbon dioxide. It also covers activation conditions, regeneration, and structural trade-offs.

\paragraph{Without-terminology answer (score: 0.84375).}
Despite the generalized terms, the answer still discusses sodium hydroxide, zeolites, and clinoptilolite, while also considering other bases and mineral adsorbents. It addresses surface basicity, pore accessibility, gas selectivity, humidity, and operating trade-offs in biogas upgrading. In this pair, rubric coverage increases from 0.59375 to 0.84375. Thus, generalizing technical terms does not necessarily lower the score when the answer continues to address the original materials and process; this single example does not establish that generalization systematically improves answer quality.

\section{Additional Results}
\label{app:search-behavior}

\subsection{Answer Quality}
\label{app:answer-quality}

Table~\ref{tab:outcome} reports the mean rubric-based answer quality behind Figure~\ref{fig:quality}. 

\begin{table*}[t]
\centering
\small
\setlength{\tabcolsep}{5pt}
\begin{tabular}{ll ccccc c}
\toprule
Benchmark & Agent & $q$ & $\qP$ & $\qS$ & $\qT$ & $\qU$ & Rel.\ change $\qU$ \\
\midrule
\multirow{3}{*}{\rqa{} }
 & \tulu{} & .791 & .694\sig{ns} & .615\sig{***} & .565\sig{***} & .294\sig{***} & $-$62.8\% \\
 & \tulugpt{} & .610 & .496\sig{*} & .466\sig{***} & .449\sig{***} & .215\sig{***} & $-$64.8\% \\
 & \tongyi{} & .840 & .763\sig{**} & .681\sig{***} & .624\sig{***} & .334\sig{***} & $-$60.2\% \\
\midrule
\multirow{3}{*}{\sqa{} }
 & \tulu{} & .477 & .504\sig{ns} & .431\sig{ns} & .438\sig{**} & .308\sig{***} & $-$35.5\% \\
 & \tulugpt{} & .524 & .495\sig{ns} & .489\sig{ns} & .496\sig{ns} & .377\sig{***} & $-$25.8\% \\
 & \tongyi{} & .438 & .414\sig{ns} & .357\sig{**} & .395\sig{ns} & .207\sig{***} & $-$52.8\% \\
\midrule
\multirow{3}{*}{\drb{}}
 & \tulu{} & .457 & .441\sig{***} & .407\sig{***} & .358\sig{***} & .258\sig{***} & $-$43.6\% \\
 & \tulugpt{} & .391 & .387\sig{ns} & .359\sig{**} & .325\sig{**} & .214\sig{***} & $-$45.4\% \\
 & \tongyi{} & .431 & .423\sig{*} & .390\sig{***} & .350\sig{***} & .259\sig{***} & $-$40.0\% \\
\bottomrule
\end{tabular}
\caption{Average answer quality for different types of query variants. Significance markers indicate differences from the fully specified condition ($q$). The last column reports the relative change from $q$ to the fully underspecified condition ($\qU$), calculated as $(\qU-q)/q$.}
\label{tab:outcome}
\end{table*}

\subsection{Clarification Behavior}
\label{app:clarify}

Table~\ref{tab:clarify-full} reports the share of Ask, Assume, and Unaware responses for each agent, benchmark, and variant behind Figure~\ref{fig:clarify}.

\begin{table*}[t]
\centering
\small
\setlength{\tabcolsep}{4pt}
\begin{tabular}{ll ccc ccc ccc}
\toprule
& & \multicolumn{3}{c}{ResearchQA} & \multicolumn{3}{c}{ScholarQA-Bench} & \multicolumn{3}{c}{DeepResearch Bench} \\
\cmidrule(lr){3-5}\cmidrule(lr){6-8}\cmidrule(lr){9-11}
Agent & Variant & Ask & Assume & Unaware & Ask & Assume & Unaware & Ask & Assume & Unaware \\
\midrule
 \multirow{4}{*}{\tulu{}} & $q_{\textsc{p}}$ & 0.0 & 11.1 & 88.9 & 0.0 & 18.4 & 81.6 & 0.0 & 37.9 & 62.1 \\
  & $q_{\textsc{s}}$ & 0.0 & 6.6 & 93.4 & 0.0 & 10.1 & 89.9 & 0.0 & 41.7 & 58.3 \\
  & $q_{\textsc{t}}$ & 0.4 & 15.5 & 84.0 & 0.0 & 15.1 & 84.9 & 0.0 & 53.3 & 46.7 \\
  & $q_{\textsc{u}}$ & 0.2 & 12.6 & 87.2 & 1.0 & 12.9 & 86.1 & 1.0 & 51.0 & 48.0 \\
\midrule
 \multirow{4}{*}{\tulugpt{}} & $q_{\textsc{p}}$ & 0.0 & 8.7 & 91.3 & 2.1 & 12.8 & 85.1 & 1.9 & 36.5 & 61.5 \\
  & $q_{\textsc{s}}$ & 1.7 & 5.2 & 93.1 & 0.0 & 13.3 & 86.7 & 0.0 & 42.9 & 57.1 \\
  & $q_{\textsc{t}}$ & 9.5 & 10.1 & 80.4 & 0.0 & 18.1 & 81.9 & 9.5 & 42.9 & 47.6 \\
  & $q_{\textsc{u}}$ & 18.1 & 7.6 & 74.3 & 12.9 & 13.9 & 73.1 & 12.6 & 32.6 & 54.7 \\
\midrule
 \multirow{4}{*}{\tongyi{}} & $q_{\textsc{p}}$ & 0.0 & 3.7 & 96.3 & 0.0 & 8.2 & 91.8 & 0.0 & 13.8 & 86.2 \\
  & $q_{\textsc{s}}$ & 0.0 & 2.6 & 97.4 & 0.0 & 3.0 & 97.0 & 0.0 & 15.3 & 84.7 \\
  & $q_{\textsc{t}}$ & 0.4 & 6.7 & 92.9 & 0.0 & 3.5 & 96.5 & 4.4 & 17.8 & 77.8 \\
  & $q_{\textsc{u}}$ & 1.2 & 6.0 & 92.8 & 0.5 & 15.5 & 84.0 & 2.0 & 18.0 & 80.0 \\
\bottomrule
\end{tabular}
\caption{Clarification behavior (\% of runs): asks for clarification (Ask), recognizes the ambiguity and states an assumption (Assume), or proceeds without recognizing the ambiguity (Unaware).}
\label{tab:clarify-full}
\end{table*}

\subsection{Query Drift}
\label{app:query-drift}

\begin{figure*}[!b]
\centering
\includegraphics[width=\textwidth]{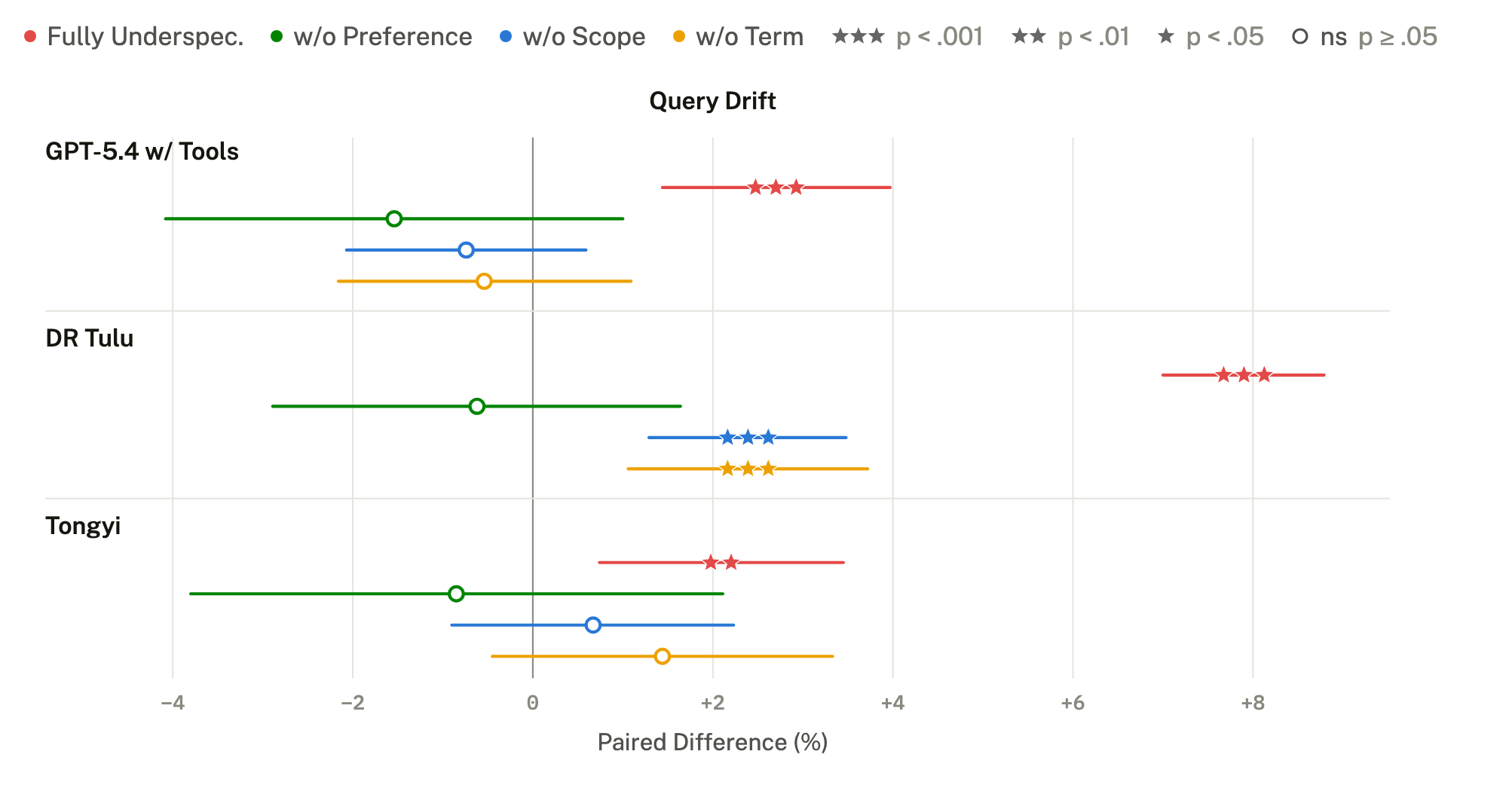}
\caption{Paired changes in query drift between the first and final search queries (underspecified minus complete query), with 95\% confidence intervals. Positive values indicate greater drift under underspecification. One, two, and three stars denote $p<.05$, $p<.01$, and $p<.001$, respectively; open circles indicate nonsignificant differences.}
\label{fig:query-drift}
\end{figure*}

\textbf{Search-query drift increases consistently across agents under severe underspecification.}\enspace Figure~\ref{fig:query-drift} shows that significant increases in first-to-final search-query drift occur across all three agents only under $\qU$, the most underspecified variant. The increase is largest for \tulu{}. By contrast, removing scope or obscuring terminology significantly increases drift only for \tulu{}, while removing preferences has no significant effect for any agent. This metric captures semantic displacement between the first and last search queries, rather than distance from the user's original information need.

\subsection{Document Utilization}
\label{app:doc-util}

Table~\ref{tab:doc-util-means} reports the mean document utilization behind the paired changes in Table~\ref{tab:citation-precision-diff}. Not every query has every type of variant, since a variant exists only if the query contains the targeted information. Each comparison therefore uses only the questions that have the given variant and were answered under both the complete query and that variant, so the complete-query mean $q$ differs slightly across variants.
\subsection{Citation Precision}
\label{app:cite-prec}

Table~\ref{tab:cite-prec-means} reports the mean citation precision behind the paired changes in Table~\ref{tab:citation-precision-diff}. As for document utilization, not every query has every type of variant, since a variant exists only if the query contains the targeted information. Each comparison therefore uses only the questions that have the given variant and were answered under both the complete query and that variant, so the complete-query mean $q$ differs slightly across variants.

\begin{table*}[!t]
\centering
\small
\setlength{\tabcolsep}{5pt}
\begin{tabular}{l cc cc cc cc}
\toprule
& \multicolumn{2}{c}{$\qP$} & \multicolumn{2}{c}{$\qS$} & \multicolumn{2}{c}{$\qT$} & \multicolumn{2}{c}{$\qU$} \\
\cmidrule(lr){2-3}\cmidrule(lr){4-5}\cmidrule(lr){6-7}\cmidrule(lr){8-9}
Agent & $q$ & $\qP$ & $q$ & $\qS$ & $q$ & $\qT$ & $q$ & $\qU$ \\
\midrule
\tulu{} & 56.39 & 56.40\sig{ns} & 55.97 & 57.48\sig{*} & 56.43 & 56.50\sig{ns} & 56.59 & 59.44\sig{***} \\
\tulugpt{} & 21.66 & 23.64\sig{ns} & 20.82 & 22.93\sig{**} & 21.29 & 20.85\sig{ns} & 21.38 & 20.94\sig{ns} \\
\midrule
Both agents & 42.36 & 43.16\sig{ns} & 41.12 & 42.88\sig{***} & 42.27 & 42.13\sig{ns} & 44.23 & 45.92\sig{***} \\
\bottomrule
\end{tabular}
\caption{Mean document utilization (\%, $\#_{\mathrm{cited}} / \#_{\mathrm{retrieved}}$) for each underspecified variant and its paired complete query $q$, pooled across \rqa{}, \sqa{}, and \drb{}. \emph{Both agents} pools the paired questions of \tulu{} and \tulugpt{}. Significance markers come from paired Wilcoxon signed-rank tests against $q$.}
\label{tab:doc-util-means}
\end{table*}

\begin{table*}[!t]
\centering
\small
\setlength{\tabcolsep}{5pt}
\begin{tabular}{l cc cc cc cc}
\toprule
& \multicolumn{2}{c}{$\qP$} & \multicolumn{2}{c}{$\qS$} & \multicolumn{2}{c}{$\qT$} & \multicolumn{2}{c}{$\qU$} \\
\cmidrule(lr){2-3}\cmidrule(lr){4-5}\cmidrule(lr){6-7}\cmidrule(lr){8-9}
Agent & $q$ & $\qP$ & $q$ & $\qS$ & $q$ & $\qT$ & $q$ & $\qU$ \\
\midrule
\tulu{} & 43.95 & 45.35\sig{ns} & 45.43 & 42.54\sig{ns} & 50.87 & 53.04\sig{ns} & 50.45 & 50.90\sig{ns} \\
\tulugpt{} & 70.52 & 75.20\sig{ns} & 70.59 & 70.11\sig{ns} & 75.78 & 76.61\sig{ns} & 70.93 & 68.79\sig{ns} \\
\bottomrule
\end{tabular}
\caption{Mean citation precision (\%, $\#_{\mathrm{supporting}} / \#_{\mathrm{cited}}$) for each underspecified variant and its paired complete query $q$, pooled across \sqa{} and \drb{}. Only questions with a final answer and at least one cited document under both conditions are included. Significance markers come from paired Wilcoxon signed-rank tests against $q$.}
\label{tab:cite-prec-means}
\end{table*}

\end{document}